# K-Syn: K-space Data Synthesis in Ultra Low-data Regimes

Yu Guan, Jianhua Zhang, Dong Liang, *Senior Member, IEEE*, Qiegen Liu, *Senior Member, IEEE*

*Abstract*—Owing to the inherently dynamic and complex characteristics of cardiac magnetic resonance (CMR) imaging, high-quality and diverse k-space data are rarely available in practice, which in turn hampers robust reconstruction of cardiac MRI. To address this challenge, we perform feature-level learning directly in the frequency domain and employ a temporal-fusion strategy as the generative guidance to synthesize k-space data. Specifically, leveraging the global representation capacity of the Fourier transform, the frequency domain can be considered as a natural global feature space. Therefore, unlike traditional methods that use pixel-level convolution for feature learning and modeling in the image domain, this letter focuses on feature-level modeling in the frequency domain, enabling stable and rich generation even with ultra low-data regimes. Moreover, leveraging the advantages of feature-level modeling in the frequency domain, we integrate k-space data across time frames with multiple fusion strategies to steer and further optimize the generative trajectory. Experimental results demonstrate that the proposed method possesses strong generative ability in ultra low-data regimes, indicating practical potential to alleviate data scarcity in cardiac MRI reconstruction.

## I. INTRODUCTION

CARDIAC magnetic resonance (CMR) imaging has emerged as a crucial imaging technique for non-invasive clinical diagnosis, due to its advantages in quantitative assessment of cardiac morphology and myocardial tissue characteristics [1]. In practice, it requires high temporal resolution to resolve cardiac motion while preserving sufficient structural detail. However, constraints on scan time and motion artifacts typically necessitate under-sampling across the entire signal space, making fully-sampled and high-quality k-space data acquisitions extremely challenging [2]. Recent breakthroughs in generative artificial intelligence (AI) have driven substantial advancements across various domains and applications [3]-[5]. Specifically, it has significantly contributed to progress in medical imaging and healthcare, while also providing a promising solution to the problem of data scarcity.

This work was supported in part by the National Key Research and Development Program of China under Grant 2023YFF1204300 and Grant 2023YFF1204302. (Corresponding author: Q. Liu)

Y. Guan is with School of Advanced Manufacturing, Nanchang University, Nanchang 330031, China. (guanyu@ncu.edu.cn)
D. Liang is with Research Center for Medical AI, Shenzhen Institutes of Advanced Technology, Chinese Academy of Sciences, Shenzhen 518055, China. (dong.liang@siat.ac.cn)
J. Zhang and Q. Liu are with School of Information Engineering, Nanchang University, Nanchang 330031, China. ({zhangjianhua@email.ncu.edu.cn, liuqiegen@ncu.edu.cn})

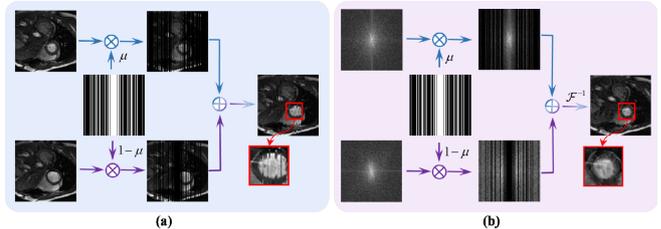

**Fig. 1.** Illustration of the difference between (a) pixel-level fusion and (b) feature-level fusion. Typically, the frequency signal is regarded as a special feature descriptor used for fusion.

The rapid growth of generative AI has been propelled by several complementary generative model families [6], including variational autoencoders (VAEs) [7], normalizing flows [8], generative adversarial networks (GANs) [9], and denoising diffusion probabilistic models (DDPMs) [10]. Each paradigm trades off sample fidelity, diversity, and training stability differently. For example, VAEs and normalizing flows are likelihood-based models, which explicitly model the data distribution and are theoretically well-founded. However, they often produce relatively modest generative results. On the contrary, GANs employ an adversarial training approach to learn the underlying probability distribution of the data, which yield sharp samples but can suffer from mode collapse and training instability [11]. DDPM addresses these limitations by utilizing a diffusion process, which adds Gaussian noise to data step by step and then learns to reverse this process to generate samples [12]. This method ensures stable training and an interpretable generative mechanism, yielding high-quality and diverse outputs. Recently, large-scale models like latent diffusion models (LDMs) [13] are widely applied in natural image domains and enabling high-fidelity synthesis.

Despite recent advancements, several obstacles continue to impede the broader application of generative models in medical imaging, particularly for CMR imaging. First, optimizing powerful generative models typically relies on large-scale and well-curated realistic training datasets—an assumption that is difficult and costly to satisfy in practical clinical settings [14]. Second, most existing generative efforts focus on producing medical images [15]-[18]. Even when multi-modal outputs are considered, raw data (i.e., k-space data) are rarely synthesized. As a result, there is still no general framework for k-space data synthesis directly, which severely restricts the generalization of AI models to real CMR imaging scenarios. Therefore, these gaps motivate a new framework that addresses two significant but challenging problems in generative model-assisted CMR imaging: (i) How to decrease the dependence on large-scale realistic data while maintaining generative fidelity and diversity? (ii) How to directly synthesis k-space data so that the outputs remain physically consistent with MR acquisition?

Fourier transform endows the frequency domain with a global receptive field, which is a natural feature space for small-sample learning. Hence, feature-level learning in fre-quency domain is an emerging and promising paradigm for k-space data synthesis, which is expected to address the aforementioned issues [19], [20]. In this letter, we propose a generative framework that performs feature-level learning in the frequency domain and leverages temporal-fusion guidance to exploit inter-frame redundancy, thereby enabling large-scale and diverse k-space data synthesis in ultra low-data regimes. As shown in Fig.1, unlike conventional methods relying on pixel-level convolution in the image domain for the estimation of prior, this paradigm revolves around the concept of feature-level fusion to extract and combine relevant features, making generative process more scalable and explainable. Remarkably, cardiac cine data exhibit strong temporal similarity that can be fused via feature-level learning to guide the generative trajectory toward being anatomically consistent and artifact-resistant. This innovative mechanism addresses the twin challenges of data scarcity and data fidelity, providing a practical path to expand downstream tasks of CMR imaging when fully-sampled acquisition is infeasible.

## II. Method

Motivated by the scarcity of high-quality k-space in dynamic CMR imaging, we propose a frequency-domain generative framework that reduces the dependence on large datasets and maintains tractable for high-resolution cine. Concretely, the framework consists of three key components: (i) temporal-fusion strategy to exploit inter-frame redundancy and enrich frequency features without additional acquisitions; (ii) a compact latent representation of k-space volumes that makes diffusion process efficient and stable; (iii) latent diffusion with temporal-fusion guidance, which steers sampling toward a structure-preserving subspace, yielding diverse and high-fidelity k-space data.

### A. Temporal-fusion Strategy in Frequency Domain

**Fourier Feature Space:** The convolution theorem states that a spatial convolution becomes point wise multiplication after a Fourier transform [21]. Hence each frequency component combines information from the entire image. Frequency domain therefore forms a feature space with an inherent global receptive field, which is advantageous when only a few training datasets are available. Specifically, the Fourier convolution transforms images in the neural network to the frequency domain and is formulated as:

$$k = \mathcal{F}(x)(u,v) = \sum_{h=0}^{H}\sum_{w=0}^{W} x(h,w) e^{-j2\pi(\frac{h}{H}u+\frac{w}{W}v)} \quad (1)$$

where $x$ and $k$ share the same spatial dimension of $H \times W$. Then, the amplitude and phase components can be respectively formulated as:

$$k^A(u,v) = \left[\mathcal{R}^2(k)(u,v) + \mathcal{I}^2(k)(u,v)\right]^{\frac{1}{2}}$$
$$k^P(u,v) = \arctan\left[\frac{\mathcal{I}(k)(u,v)}{\mathcal{R}(k)(u,v)}\right] \quad (2)$$

where $\mathcal{R}(k)$ and $\mathcal{I}(k)$ represent the real and imaginary part respectively. Since $k^A(u,v)$ and $k^P(u,v)$ normally reflect style-relevant information and structural-relevant information respectively, it is intuitive to generate new styles by simply manipulating on $k^A(u,v)$ while remaining $k^P(u,v)$ constant [22]. Therefore, random recombination of the amplitude components in the frequency domain，while keeping the phase components intact to preserve anatomical fidelity, makes it possible to generate vast numbers of diverse synthetic samples from only limited real data.

**Temporal-fusion Guidance:** The unique characteristic of CMR imaging exhibits a high degree of correlation between different frames. Based on the advantages of feature-level space and the distinctive properties of temporal redundancy, we fuse adjacent frames at the frequency domain by concatenating their amplitude components and projecting them into a fused frequency signal. As shown in Fig. 2, the temporal fusion process can be expressed mathematically as follows:

$$K^A(t) = \mu k_m^A(t_m) + (1-\mu) k_n^A(t_n) \quad (3)$$

where $t_m$ and $t_n$ denote distinct time points in ECG-gated segmented imaging, $k_m^A$ and $k_n^A$ denote the frames acquired at the corresponding time points. Given a set of random $\mu \sim U(0,1)$ values, the temporal-fusion can provide various k-space data styles. By inheriting useful energy from multiple frames and retaining at least as many independent frequency components as any single frame, the fused signal guides the generative trajectory toward a feasible subspace that is structure-preserving and spectrally consistent, thereby enabling effective learning even from limited data.

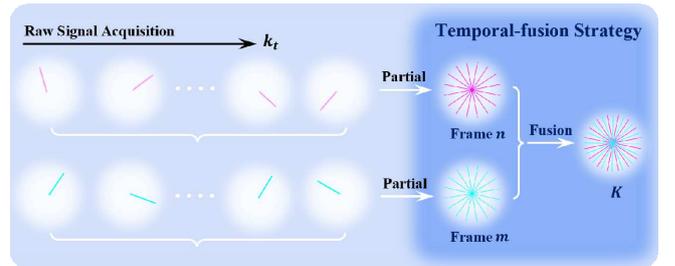

**Fig. 2.** Temporal-fusion strategy. Partial per-frame k-space data is fused across adjacent frames to form different frequency feature.

### B. Latent Representation of K-space Data

To enable efficient k-space data generation, we develop a novel compression framework that embeds high-resolution multi-frame k-space volumes into a compact and semantically meaningful latent space. Concretely, a frequency encoder $\mathcal{E}_\theta(\bullet)$ operates on overlapping k-space patches using a two-channel representation of the complex signal and lightweight convolutions to capture local structure. Afterwards, these quantized features are aggregated by a shallow volume module to form a global latent tensor $z$ with markedly reduced spatial-temporal resolution:

$$z = \mathcal{E}_\theta(K) \quad (4)$$

Following this, the joint decoder $\mathcal{D}_\theta(\bullet)$ produces the final reconstruction from the quantized features as follows:

$$\tilde{K} = \mathcal{D}_\theta(z) \quad (5)$$

Generative modeling of latent representations with our trained frequency compression models consisting of encoder $\mathcal{E}_\theta(\bullet)$ and decoder $\mathcal{D}_\theta(\bullet)$, we now have access to an efficient and low-dimensional latent space in which high-frequency and imperceptible details are abstracted away [23]. Compared to the high-dimensional pixel-level space, this latent space is more suitable for likelihood-based generative models, as they can train in a lower dimensional and computationally much more efficient space.

## C. Latent Diffusion with Temporal-fusion Guidance

Following compression, we perform temporal-fusion guided generation in this latent diffusion space. Fig. 3 illustrates the latent diffusion pipeline with temporal-fusion guidance. Specifically, dynamic k-space frames are fused in the frequency domain using multiple schemes—adjacent-frame, skip-frame and grouped fusion—to construct a guidance frequency signal. Each scheme applies frequency-dependent weighting to aggregate complementary information across time. As mentioned above, we have encoded the fused frequency signal into latent features $z = \varepsilon_\theta(K)$. Therefore, diffusion model first defines a forward noising process that gradually corrupts latent feature $z_0$ into $z_t$ over $T$ time steps:

$$q(z_t | z_0) = \mathcal{N}(z_t; \sqrt{\bar{\alpha}_t} z_0, (1-\bar{\alpha}_t)\mathbf{I}) \quad (6)$$

where $\sqrt{\bar{\alpha}_t} z_0$ is the mean and $(1-\bar{\alpha}_t)\mathbf{I}$ is the covariance of the distribution at timestep $t$, with $T$ denoting the total number of timesteps. $\mathbf{I}$ is the identity matrix and $\alpha_t \in (0,1)$ is a noise-level hyperparameter. During training, the model learns to predict the noise $\epsilon$ added at each step via a conditional U-Net denoiser $\epsilon_\theta$, using both the noised latent feature $z_t$ and the condition embedding $c$ as inputs:

$$\mathcal{L}_{diff} = \mathbb{E}_{z_0,t,c,\epsilon}[\| \epsilon - \epsilon_\theta(z_t,t,c) \|^2] \quad (7)$$

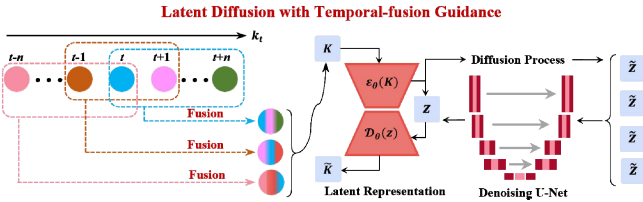

**Fig. 3.** Schematic of different temporal-fusion strategies (adjacent, skip, and grouped) jointly guiding latent diffusion process.

Once the model is trained, we generate samples by reverse latent diffusion process, starting from $z_T \sim \mathcal{N}(0,\mathbf{I})$ and iteratively applying:

$$z_{t-1} = (\frac{1}{\sqrt{\alpha_t}})(z_t - (\frac{1-\alpha_t}{\sqrt{1-\bar{\alpha}_t}})\epsilon_\theta(z_t,t,c)) + \sigma_t \cdot \eta_t \quad (8)$$

where $\eta_t \sim \mathcal{N}(0,\mathbf{I})$ and $\sigma_t$ controls the stochasticity. Importantly, temporal-fusion features encoded as $z$ are injected into the model at every scale and time step to guide the diffusion process. This step-wise conditioning draws the Markov chain toward a well-constrained subset of latent states, enabling faster and more efficient k-space synthesis.

## III. EXPERIMENTS

### A. Experimental Setup

*Datasets*: The fully-sampled cardiac cine data used in the experiment are provided by the Shenzhen Institutes of Advanced Technology, Chinese Academy of Sciences. Specifically, all data are collected on a 3.0T scanner (SIEMENS EQUIPTOM Trio) equipped with a 20-channel coil using a balanced steady-state free precession (BSSFP) sequence from 30 volunteers. The following sequence parameters are used: FOV = 330×330 mm², acquisition matrix = 256×256, slice thickness = 6 mm, TR/TE = 3.0 ms/1.5 ms. The acquired temporal resolution is 40.0 ms. Notably, we utilize the adaptive coil combination method to combine the original multi-coil data to generate single-coil complex-valued data. Finally, we directly employ 2D-t cardiac MR data of size 192×192×16 ($x \times y \times k_t$) for training.

*Evaluation Metrics*: We evaluate the fidelity and diversity of synthetic k-space data using Fréchet Inception Distance (FID), Kernel Inception Distance (KID) and Squared Maximum Mean Discrepancy (MMD²), which assess the similarity between the distributions of synthetic data and real data, with lower values indicating higher synthesis quality. Besides, we utilize Peak Signal-to-noise Ratio (PSNR), Structural Similarity Index (SSIM) and Mean Squared Error (MSE) for quantitative evaluation in the reconstruction experiment.

*Implementation*: We employ a two-stage optimization scheme for the proposed K-Syn model. In stage-1, the encoder-decoder is optimized on 64×64×3 patches for 350 iterations with a learning rate of 4.5×10⁻⁶. In stage-2, the latent diffusion model is trained with an $L_1$ objective under a cosine noise schedule for 200 iterations at a learning rate of 1.0×10⁻⁶. We utilize the Adam optimizer and maintain an exponential moving average of the parameters to promote stable and efficient training. For other synthesis models, we utilized their github repositories and adhered to the specified original hyperparameters. All experiments are implemented on an Ubuntu 20.04 LTS (64-bit) operating system equipped with an Intel i7 (CPU) and NVIDIA 4090 (GPU, 24GB memory) in the open framework PyTorch with CUDA and CUDNN support. Open-source code related to this letter is available at https://github.com/yqx7150/K-Syn.

TABLE I
QUANTITATIVE RESULTS OF K-SYN AGAINST STATE-OF-THE-ART GENERATIVE METHODS.

| Method | FID ↓ | KID ↓ | MMD² ↓ |
|---|---|---|---|
| GAN-based | 137.2536 | 0.0370 ± 0.0008 | 0.2126 |
| DM-based | 96.7970 | 0.0172 ± 0.0008 | 0.0923 |
| Ours (K-Syn) | 31.9902 | 0.0051 ± 0.0004 | 0.0329 |

### B. Evaluation of Synthesis Quality

We assess the synthesis quality of the proposed K-Syn framework by conducting comparisons with several state-of-the-art generative methods, including GANs and DMs. Specifically, we train all models on fully-sampled cardiac MR data and generate 2000 synthetic samples for evaluation. The quantitative experimental results are provided in Table I, which shows that K-Syn achieves the lowest FID, KID, and MMD² scores, indicating superior generative fidelity and realism. Compared to the second-best method, FID and KID scores for K-Syn are over two times lower, highlighting a significant improvement in generative fidelity.

Regular visual assessment tests are conducted throughout the experimentation phase and Fig. 4 presents representative k-space data generated by different methods along with their corresponding image-domain results. The cardiac region is magnified to highlight differences in anatomical detail generation. It is evident that GAN-based method predominantly generates coarse structural outlines in the image domain, reflecting a pronounced limitation in its ability to model the complex distribution of k-space data. DM-based method demonstrates improved synthesis of low-frequency details, resulting in more accurate image-domain contours and contrast. However, its high-frequency synthesis remains limited and the lack of high-frequency generation leads to smoother textures and a loss of anatomical features. In contrast, K-Syn effectively captures both low- and high-frequency components, resulting in high-resolution generation with well-preserved myocardial boundaries and reduced artifacts.

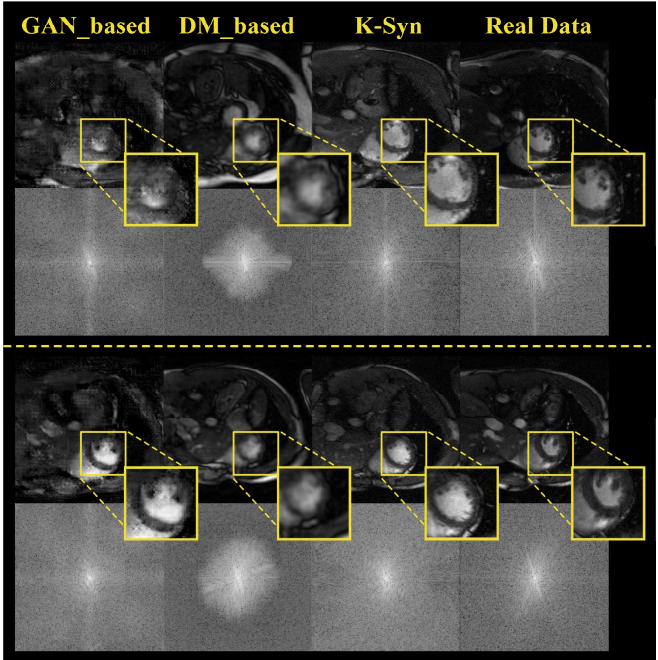

**Fig. 4.** Qualitative comparison of overall quality in cardiac k-space data synthesis. K-Syn yields sharper myocardial detail and fewer artifacts.

### C. K-space Data Synthesis in Ultra Low-Data Regimes

To evaluate the generative ability of K-Syn in ultra low-data regimes, we conduct a comparative experiment through the following training data settings: (1) Real-50/200: Partial data are selected from the real data using principal component analysis (PCA). (2) K-Syn-50/200: Partial data are selected from the real data using PCA and processed using the temporal-fusion strategy, which exploiting interframe redundancy of cardiac data and enrich frequency features. All models are trained under the same experimental configuration and the overall quantitative results are presented in Table II. K-Syn-50/200 achieves lower FID and KID scores compared with Real-50/200, indicating the effectiveness of the temporal-fusion strategy. Notably, the quantitative performance of K-Syn-50 is comparable to Real-200. Fig. 5 shows that the K-Syn-50 preserves both low and high frequency structures in k-space data as effectively as Real-200, highlighting the remarkable generative capability of K-Syn in ultra low-data regimes.

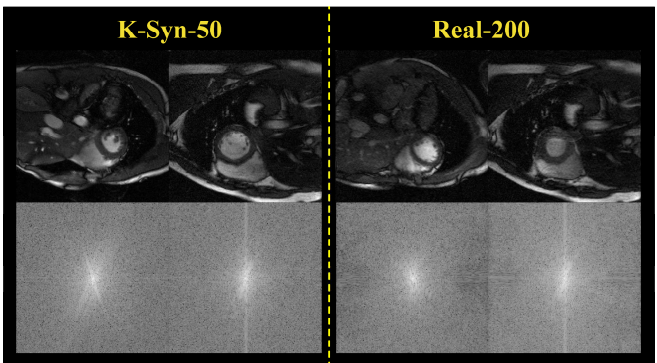

**Fig. 5.** Performance comparison of cardiac k-space data synthesis results generated by K-Syn model trained with different numbers of samples.

TABLE II
QUANTITATIVE RESULTS OF K-SYN MODELS TRAINED WITH DIFFERENT DATASET SIZES.

| Method | FID ↓ | KID ↓ | MMD² ↓ |
|---|---|---|---|
| Real-50 | 38.1126 | 0.0061 ± 0.0004 | 0.0798 |
| Real-200 | 31.9902 | 0.0051 ± 0.0004 | 0.0300 |
| K-Syn-50 | 28.3000 | 0.0022 ± 0.0003 | 0.0100 |
| K-Syn-200 | 33.7968 | 0.0055 ± 0.0004 | 0.0418 |

### D. Synthetic K-space Data for Downstream Task

In this experiment, we aim to accelerate MRI reconstruction by leveraging a representative reconstruction model WKGM to map the under-sampled k-space data to the fully-sampled reconstructions. Specifically, we conduct a comparative experiment with two models trained separately on synthetic cardiac data (WKGM-Syn) and real cardiac data (WKGM) to assess the validity of synthetic k-space data. The visualization results of different methods at 10-fold acceleration are shown in Fig. 6. One can see that WKGM-Syn attains quantitative performance comparable to that of WKGM. It is worth noting that WKGM-Syn is trained solely on synthetic data yet demonstrates strong reconstruction performance, which indicates that the synthetic data effectively capture the essential characteristics of real k-space data. Moreover, experimental results demonstrate the feasibility of leveraging the powerful generative prior knowledge learned by K-Syn to facilitate MRI reconstruction.

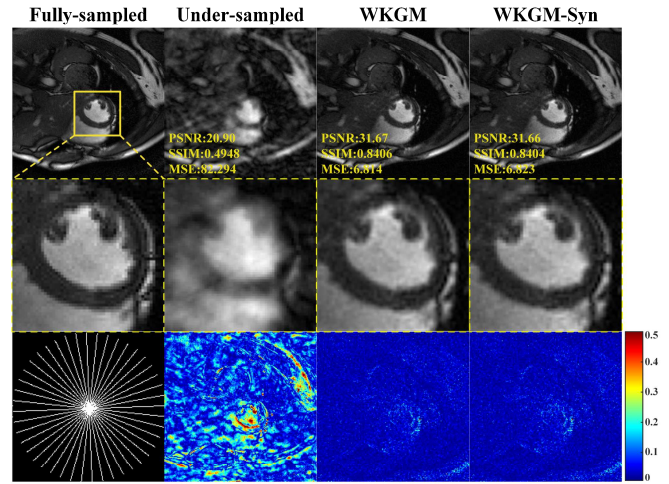

**Fig. 6.** Reconstruction of the cardiac data at radial sampling of $R$=10. The second row shows the enlarged view of the ROI region and the third row shows the error map of the reconstruction.

## IV. CONCLUSION

In conclusion, this letter tackles a central bottleneck in dynamic CMR imaging—the scarcity of high-quality and diverse k-space data—by shifting generative modeling from the image-domain to the frequency-domain. By treating the frequency-domain as a natural global feature space afforded by the Fourier transform, we perform feature-level learning and fuse different time frames to enrich frequency cues without additional acquisitions. This design stabilizes and strengthens generation in ultra low-data regimes and steers the sampling trajectory toward a structure-preserving and data consistent subspace. Experimental results indicate the practical potential to mitigate data scarcity and support downstream reconstruction workflows where physical consistency with raw measurements matters.